\pdfoutput=1

\documentclass[11pt]{article}

\usepackage{acl}

\usepackage{times}
\usepackage{latexsym}

\usepackage[T1]{fontenc}

\usepackage[utf8]{inputenc}

\usepackage{microtype}

\usepackage{inconsolata}

\newcommand{\readme}{\texttt{README}}
\newcommand{\FuncRead}{\texttt{FuncRead}}

\usepackage{array}
\usepackage{bm}
\usepackage{float}

\usepackage{stfloats}

\usepackage{times}
\usepackage{latexsym}
\usepackage{hyperref}

\newcommand{\RNum}[1]{\uppercase\expandafter{\romannumeral #1\relax}}

\usepackage{multirow}
\usepackage{tcolorbox}
\usepackage{tabularx}
\usepackage{makecell}
\usepackage{amsmath,amssymb}
\usepackage{physics}
\usepackage{booktabs}
\usepackage{enumitem}
\usepackage{subcaption}
\usepackage[switch]{lineno}

\title{\textit{Read between the lines} - Functionality Extraction From \texttt{README}s}

\author{Prince Kumar, Srikanth Tamilselvam, Dinesh Garg \\
IBM Research AI \\
prince.kumar12@ibm.com, \{srikanth.tamilselvam, garg.dinesh\}@in.ibm.com
}

\begin{document}
\maketitle

\begin{abstract}
While {\em text summarization} is a well-known NLP task, in this paper, we introduce a novel and useful variant of it called {\em functionality extraction from Git {\readme{}} files}. Though this task is a {\em text2text} generation at an abstract level, it involves its own peculiarities and challenges making existing {\em text2text} generation systems not very useful. The motivation behind this task stems from a recent surge in research and development activities around the use of large language models for code-related tasks, such as code refactoring, code summarization, etc. We also release a human-annotated dataset called \FuncRead, and develop a battery of models for the task. Our exhaustive experimentation shows that small size fine-tuned models beat any baseline models that can be designed using popular black-box or white-box large language models (LLMs) such as  ChatGPT \cite{chatgpt} and Bard \cite{chowdhery2022palm}. Our best fine-tuned 7 Billion CodeLlama model exhibit $70\%$ and $20\%$ gain on the $F_1$ score against ChatGPT and Bard respectively. 

\end{abstract}
\section{Introduction}
Large Language Models (LLMs) are known to perform really well on many {\em text2text} \cite{yang2021towards} generation tasks such as {\em summarization} \cite{liu2019text, el2021automatic}), {\em translation} \cite{wang2019learning, maruf2021survey}, etc. Because of this success, there is a growing research interest in applying LLMs in novel task settings such as {\em explaining complex codes, generating new recipes, simplifying contents,} etc\footnote{https://platform.openai.com/examples}. In this paper, we introduce another novel task called {\em functionality extraction from Git {\readme} files} -- a variant of {\em text summarization} task \cite{prana2019categorizing} that detects all the functionalities supported by the corresponding application software. This task can also be seen as a variation of a Question-Answering (QA) \cite{fan2019eli5, soares2020literature} task where the question like \textit{List all functionalities} is fixed. 

\begin{figure*}[ht]
  \centering
  \includegraphics[width=0.819\textwidth]{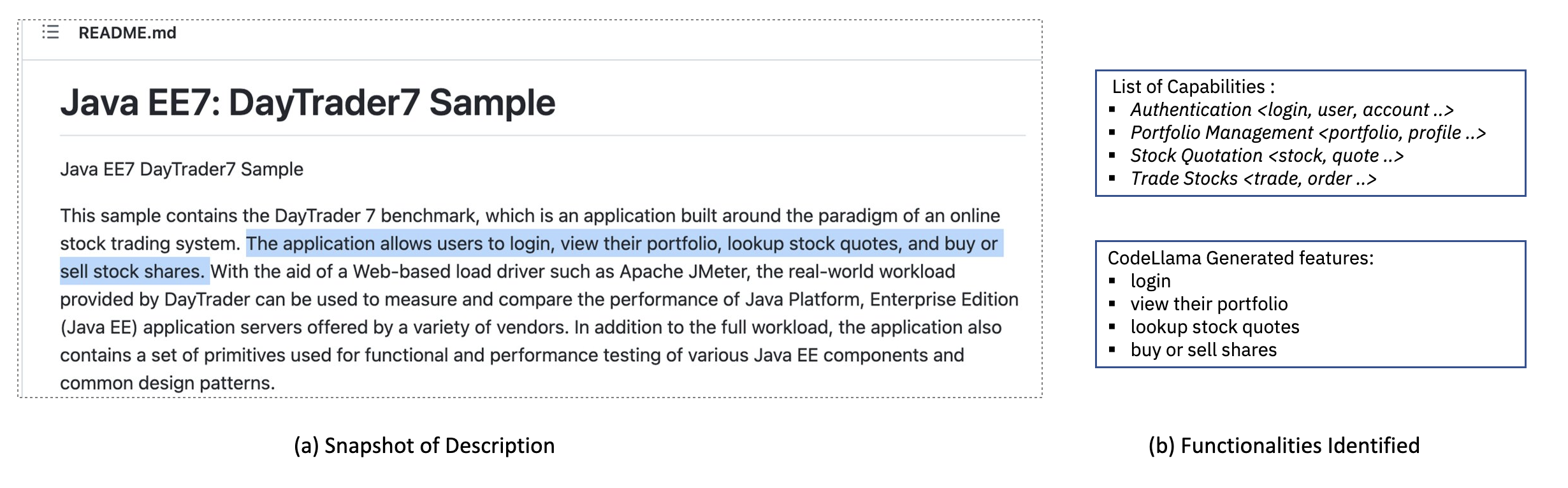}
 \caption{Snapshot of Github README content of Daytrader, an online trading application is captured in (a). The 
 human annotated four functionalities based on the description are listed as golden truth along with the functionalities generated by fine-tuned 7 billion CodeLlama model.}
  \label{fig:desc_func}
\end{figure*}

The motivation to introduce {\em automatic functionality extraction from Git {\readme} files} stems from the requirement of application code refactoring to decompose a monolith application into functional microservices. Here each microservice is a collection of closely connected application artifacts (programs, tables etc.) supporting a common functionality \cite{fowler, richardson2018microservices, newman2021building}. Current microservice recommendation systems rely a lot on subject matter experts (SMEs) and falls short to correctly group artefacts since they do not have reference list of functionalities. But many application Git {\readme} files tend to contain capture {\em different functionalities \footnote{Occasionally, we call {\em functionality} as {\em feature}} of the underlying software code base}\footnote{\url{https://docs.GitHub.com/en/repositories/managing-your-repositorys-settings-and-features/customizing-your-repository/about-readmes}} along with other implementation details like {\em what it does, how others can use it, licensing, etc.,}\cite{prana2019categorizing, chen2021evaluating}. As an example, the {\readme} file of the Daytrader application\footnote{\url{https://GitHub.com/WASdev/sample.daytrader7/}} discusses {\em the application overview, the technology used, licensing terms,} etc., and in between discusses {\em four functionalities} as highlighted in Figure \ref{fig:desc_func}(a).

Recently, \cite{doan2023too} focused on leveraging LLM to generate sections of README.md like "About" section (brief 1-2 line summary of repo) but they do not aim to list all the functionalities. Extraction of the application functionalities from such {\readme} files is not straightforward. The functionalities may not be always structured and might spread across multiple paragraphs and lines. Therefore, there is a need for an intelligent system that can parse the text, understand functionality expressions, de-duplicate, and list them. To tackle this first-of-its-kind task, we also introduce and release a new dataset called \textit{{\FuncRead} } that will help the community to benchmark their functionality understanding module and refactor monolith applications into discovered functional microservices. The key contributions of this paper are as follows.
\begin{enumerate}[leftmargin=*,noitemsep]
\item We introduce a novel {\em functionality extraction from Git {\readme} files} task and human-annotated dataset called {\FuncRead}. This dataset captures the human-annotated lists of the functionalities in both extractive and abstractive forms for each of $2101$ different GitHub {\readme} files following permissible licenses. 
\item We perform a comparative analysis of generative models to reason out the gap in performance between different baselines on the {\FuncRead} dataset. To enable comparison, we perform bipartite matching (one-to-one, many-to-one, and weighted many-to-one) to align generated functionalities with the gold functionalities.
\item We present smaller fine-tuned generative models 1\&7 billion StarCoderbase, 2.7 billion phi-2, 7 billion Llama-2 \& CodeLlama which give superior results compared to ChatGPT and Bard.
\end{enumerate}


\section{{\FuncRead} Dataset}
The {\FuncRead} dataset is a first-of-its-kind dataset that consists of functionalities described in the {\readme} files. These functionalities were hand-curated by human annotators after carefully reading the file. For each {\readme} file, the functionalities are annotated in two formats - {\em extractive} and {\em abstractive}. Extractive functionalities are segments of the text or span from the {\readme} file; whereas abstractive functionalities are the self-explained versions of the corresponding extractive functionalities, written in the annotator's own words. Each of these format outputs are presented in the form of a list. The dataset consists of unique $2101$ human annotated GitHub {\readme} files. 
\subsection{Dataset Collection}
\label{sec:dataset-collection}
We used GitHub provided APIs to randomly select a subset of public repositories that comes with a permissible licenses. Further, we manually inspected the {\readme} files of these repositories and retained only the ones that comprised of at least two functionalities. Note, we do not store the {\readme} files for the crawled repositories, we only extracted the {\readme} content and other metadata like license information. We also removed markdown tags and any Personal Identifiable Information (PII) like names, email addresses etc. before further processing. The license distribution for the $2101$ {\readme} files are as follows MIT (\em{1436}), Apache (\em{334}) , BSD (\em{334}), and EPL (\em{6})  licenses. We found that the majority of the repositories consist of $10$ or lesser functionalities with an average being $5$ functionality per repository. Some repository has as many as $34$ different functionalities.


\subsection{Dataset Annotation}
We had a total of seven annotators involved in the initial data annotation process. Each annotator was asked to read the whole {\readme} file and perform both the annotations -- {\em extractive} and {\em abstractive}. For extractive annotation, annotators were asked to select text spans from the {\readme} file which they felt were describing functionalities, and note them in the form of a numbered list. For abstractive annotation, each annotator was asked to describe the functionalities in their own words. All the annotators were given a disjoint set of {\readme} files.
\subsection{Annotation Validation}
\begin{figure}[h!]
     \centering
\includegraphics[width=\columnwidth]{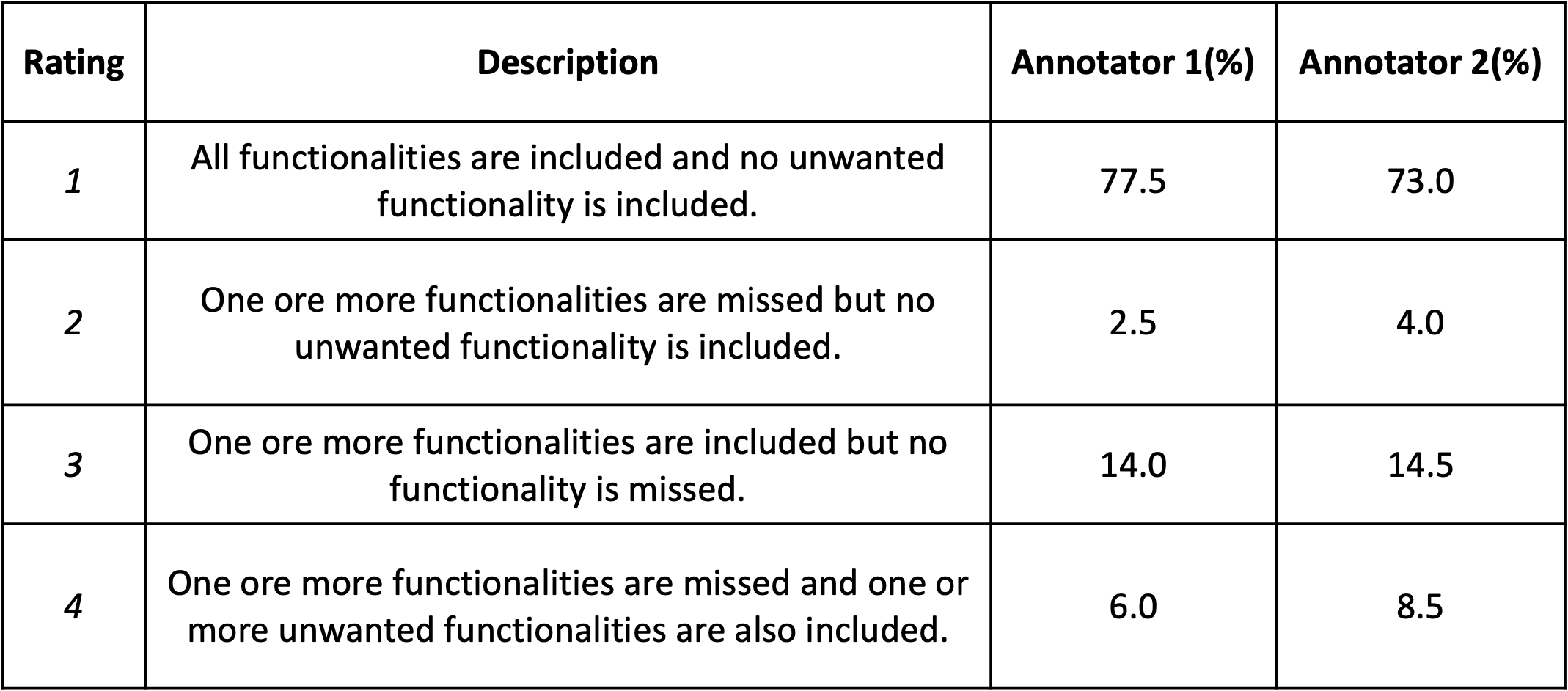}
  \caption{Ratings distribution of the two annotators during the verification step of the FuncRead dataset.}
  \label{fig:rating_matching}
\end{figure}

We employed two new independent annotators for the purpose of human validation of the dataset obtained from the previous step. We randomly sampled $200$ {\readme} files from each of these two annotators out of which $50$ {\readme} files were common for both the annotators. Both of these annotators were instructed to read extractive as well as abstractive functionalities and check whether all the functionalities were included. Based on their observation, they were tasked to give a rating from $1$ to $4$ based on the degree of strictly necessary functionalities annotated. These ratings were used to calculate the inter-annotator agreement. We observed a Kappa score of $0.873$. Figure \ref{fig:rating_matching} describes the ratings and the rating score distribution for both.

More details on the dataset characteristics and annotation procedure can be found in appendix.

 
  
		
		

\section{Task Modelling}
For modeling purposes, one can view the {\em functionality extraction} as a generation task. In the generation mode, the goal is to generate a list of functionalities from a given {\readme{}} file. 
As ours is the first-of-its-kind dataset, we used ChatGPT and Bard models known to perform really well on most NLP and code tasks even in zero-shot setting as a baseline for our task. Among many prompts, the following prompt ``{\em List all the features from above text. Each features should be in individual line without headings. Each features should be in individual line without headings. Do not include features related to license}'' provided the best results. The actual list of prompts tried on ChatGPT and Bard can be found in section \ref{sec:taskmodellinggpt}.
We wanted to study if task specific small sized models can provide competitive results. For this we considered mix of NL and code model variants like 1b and 7b StarCoderbase, 2.7b phi-2 and 7b llama-2 and CodeLlama. For fine-tuning, we pre-processed the {\readme} data through the steps listed in section \ref{sec:dataset-collection}. Next, we append it with ``\textit{\textbackslash n\#\#FEATURES\#\#\textbackslash n}'' as the task designator prompt followed by the human annotated list of functionalities corresponding to that {\readme} file. For inference, we simply appended the task designator prompt to the {\readme} text and then allowed the model to complete sequence to generate list of functionalities.

\section{Experiments and Results}

\begin{table*}[!ht]
\setlength\tabcolsep{8.5pt}
	\setlength\tabcolsep{10.5pt}
 
  
		

 \begin{tabularx}{\textwidth}{crrrrrrrr}
    \toprule
    \thead{\bfseries Model} & \thead{\textbf{$F_1^\#$}} & \thead{\bfseries $P^\#$} & \thead{\bfseries $R^\#$} & \thead{\textbf{$F_1^*$}} & \thead{\bfseries $P^*$} & \thead{\bfseries $R^*$} & \thead{\textbf{$F_1^+$}} & \thead{\bfseries $P^+$} \\ \midrule
    ChatGPT & 0.459 & 0.336 & \textbf{0.900} & 0.431 & 0.303 & \textbf{0.922} & 0.406 & 0.282 \\
    Bard & 0.653 & 0.611 & 0.806 & 0.649 & 0.573 & 0.858 & 0.612 & 0.528 \\
    StarCoderbase-1b & 0.772 & 0.816 & 0.786 & 0.808 & 0.788 & 0.876 & 0.754 & 0.711 \\
    StarCoderbase-7b & 0.743 & 0.797 & 0.754 & 0.787 & 0.777 & 0.844 & 0.734 & 0.698 \\
    Phi-2 & 0.231 & 0.172 & 0.656 & 0.226 & 0.159 & 0.733 & 0.207 & 0.144 \\
    Llama2-7b & 0.698 & 0.748 & 0.715 & 0.715 & 0.700 & 0.795 & 0.658 & 0.622 \\
    CodeLlama-7b & \textbf{0.784} & \textbf{0.827} & 0.794 & \textbf{0.816} & \textbf{0.801} & 0.877 & \textbf{0.770} & \textbf{0.738} \\
    \bottomrule
\end{tabularx}

\caption{Result comparison for various fine-tuned models against out-of-the box large models for threshold = $0.3$. \# represents one-to-one bipartite matching, * represents many-to-one bipartite matching, + represents weighted many-to-one  bipartite matching. 
}

\label{table:results_1_0.3}

\end{table*}
\begin{table*}[!ht]
	\setlength\tabcolsep{8.5pt}
	\begin{tabularx}{\textwidth}{crrrrrrrrr}
 
		\toprule
        
        \multirow{2}{*}{\thead{\bf Model}}  & \multicolumn{3}{c}{\thead{\bf ROUGE-1}} & \multicolumn{3}{c}{\thead{\bf ROUGE-2}} & \multicolumn{3}{c}{\thead{\bf ROUGE-L}} \\
        \cmidrule(l{5mm}r{5mm}){2-4}  \cmidrule(l{5mm}r{5mm}){5-7} \cmidrule(l{5mm}r{5mm}){8-10}
		  & \thead{\bm {$F_1$}} & \thead{\bf P} & \thead{\bf R} &\thead{\bm {$F_1$}}  & \thead{\bf P} & \thead{\bf R} &\thead{\bm {$F_1$}}  & \thead{\bf P} & \thead{\bf R} \\ %
		\midrule
		
        ChatGPT & 0.423 & 0.404 & 0.564 & 0.301 & 0.291 & 0.391 & 0.410 & 0.390 & 0.549	\\ 
        Bard & 0.616 & 0.648 & 0.673 & 0.511 & 0.542 & 0.549 & 0.609 & 0.640 & 0.666	 \\ 
        StarCoderbase-1b & 0.759 & 0.750 & \textbf{0.845} & 0.676 & 0.667 & \textbf{0.755} & 0.757 & 0.747 & \textbf{0.842}	\\
        StarCoderbase-7b & 0.754 & 0.790 & 0.802	 & 0.640 & 0.663 & 0.688 & 0.752	 & 0.788 & 0.800\\
        Phi-2 & 0.665 & 0.677 & 0.765 & 0.567 & 0.571 & 0.658 & 0.663 & 0.674 & 0.762 \\
        Llama2-7b & 0.755 & 0.787 & 0.810 & 0.659 & 0.688 & 0.706 & 0.752 & 0.783 & 0.806 \\
        CodeLlama-7b & \textbf{0.778} & \textbf{0.815} & 0.820 & \textbf{0.684} & \textbf{0.710} & 0.725 & \textbf{0.777} & \textbf{0.813} & 0.818 \\
        \bottomrule
	\end{tabularx}

\caption{Results for one-to-one matched pairs of different models generation and ground truth for threshold = $0.3$.
}

\label{table:results_one2one_lexical_0.3}

\end{table*}

\begin{table}[!ht]
	\setlength\tabcolsep{9pt}
	\begin{tabularx}{\textwidth}{rrrr}
 
		\cmidrule[\heavyrulewidth]{1-4}

        \multirow{2}{*}{\thead{\bf Model}}  & \multicolumn{3}{c}{\thead{\bf BERTScore}}\\
        \cmidrule(l{5mm}r{5mm}){2-4}
		   & \thead{ $\bm F_1$ } & \thead{\bf P} & \thead{\bf R} \\ %
		\cmidrule{1-4}
		
         ChatGPT & 0.895 & 0.889 & 0.902	\\ 
        Bard & 0.912 & 0.910 & 0.916 \\ 
        StarCoderbase-1b & 0.945 & 0.940 & \textbf{0.951}	\\
        StarCoderbase-7b & 0.938 & 0.938 & 0.940 \\
        Phi-2 & 0.928 & 0.925 & 0.933	\\
        Llama2-7b & 0.936 & 0.935 & 0.939 \\
		CodeLlama-7b & \textbf{0.946} & \textbf{0.946} & 0.947 \\
		\cmidrule[\heavyrulewidth]{1-4}
	\end{tabularx}

\caption{Results for one-to-one matched pairs for threshold = $0.3$.}

\label{table:results_one2one_semantic_0.3}

\end{table}

For our experiments, we divided the {\FuncRead} dataset into train, validation, and test sets comprising $1801$, $100$, and $200$ samples respectively. 
\begin{figure}[h!]
     \centering
\includegraphics[width=\columnwidth]{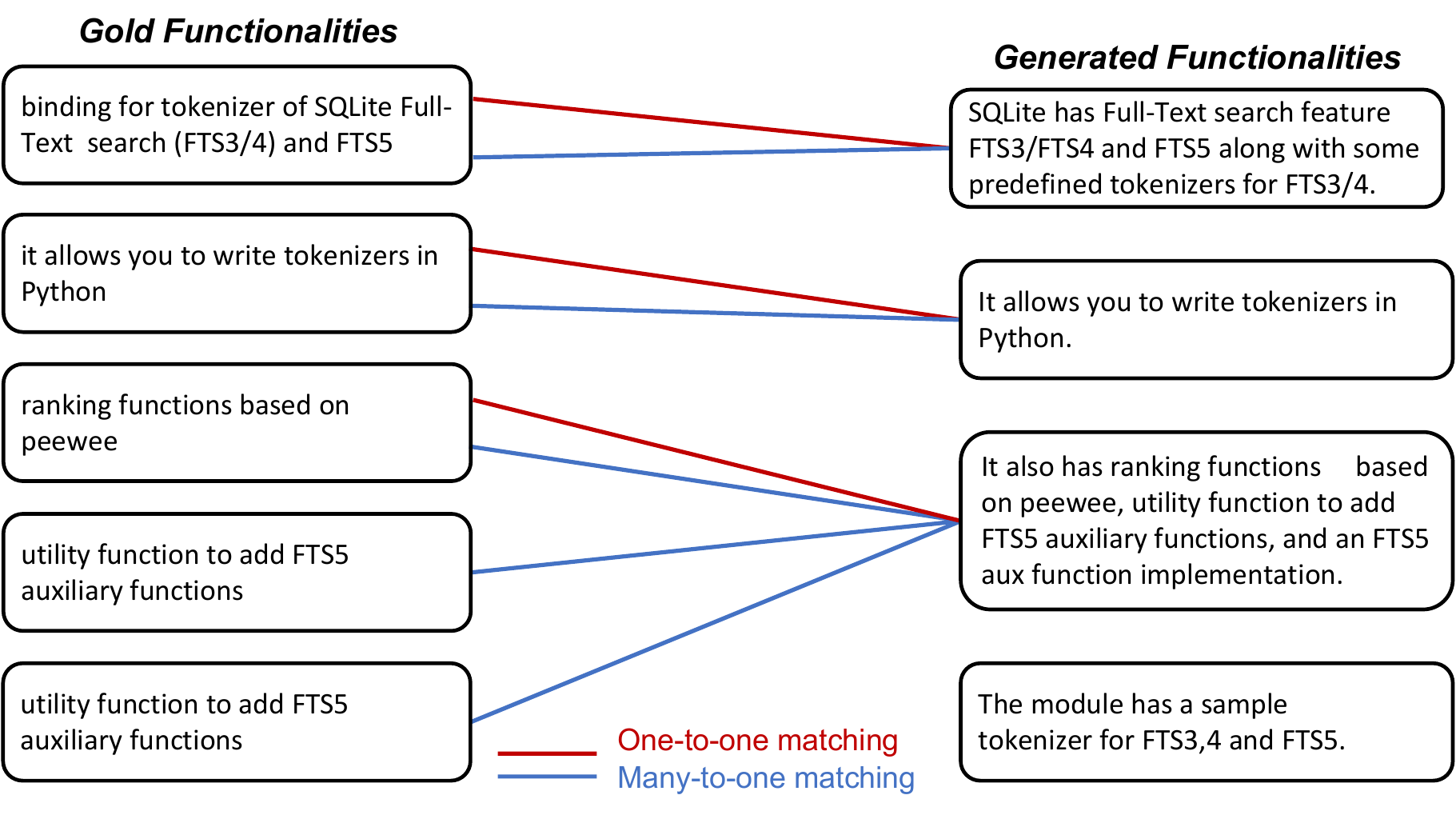}
  \caption{One-to-One bipartite matching (red color) and Many-to-one bipartite matching (blue color). Edges are established based on cosine similarity }
  \label{fig:matching}
\end{figure}

\subsection{Evaluation Metrics}
To evaluate the quality of the generated functionalities, we align them to the gold annotated functionalities via bipartite matching. We perform three kinds of bipartite matching: i) one-to-one, ii) one-to-many, and iii) weighted one-to-many. 

In any of these bipartite graphs, we have model-generated functionalities as nodes on one side and gold (ground truth) functionalities as nodes on the other side. The presence or absence of an edge in this bipartite graph is decided by the similarity scores between the corresponding sentences. In our experiments, we found threshold $0.3$ similarity matches the most with the human judgment. We did maximum bipartite matching to compute Precision ($P$), Recall ($R$), and $F_1$ scores based on matched pairs to measure the generation capability. 

For fine-tuning the models, we used extractive functionalities as gold, and because of it, we employed ROUGE-$1$, ROUGE-$2$, ROUGE-L scores to check the lexical matching quality of generated functionalities at an individual level. Since all the considered models are generative models, there is a high chance that it would introduce new tokens while generating functionalities. Hence, we also used BERTScore \cite{zhang2019bertscore} to capture the semantic similarity between the matched pairs.

\subsection{Results}
Overall, we find fine-tuned models specifically code models are reliable for this novel task. From table \ref{table:results_1_0.3}, we can observe fine-tuned models have a tendency to combine multiple functionalities into a single sentence but $F_1$, $P$, and $R$ scores of many-to-one bipartite matching indicates that it still does less frequently. But all the fine-tuned models significantly outperform ChatGPT, Bard on $P$ and $F_1$ measures. Due to inherent verbosity, $R$ is higher for the latter models. Table \ref{table:results_one2one_lexical_0.3} ROUGE scores demonstrates that the functionalities generated by the fine-tuned models have a relatively higher token similarity when matched one-to-one (it is consistent for the other two schemes as can be seen in appendix). Table \ref{table:results_one2one_semantic_0.3} BERTScores are also consistent with the claims showing better semantic similarity for the fine-tuned models. We suspect code models tendency to outperform NL models can be due to their stronger exposure to Git data. 
In few instances the models did not list any functionalities which can be attributed to complexity and lack in standardization of GitHub {\readme} files. 
Please refer to appendix for in-depth comparisons and discussions.


\section{Conclusion}
We introduced a novel task 
{\em functionality extraction from Git {\readme{}} files} and studied on a new dataset curated from public repositories to demonstrate reliability of small sized fine-tuned LLMs.


\bibliography{anthology,custom}

\clearpage

\section{Appendix}
\label{sec:appendix}
We organize the appendix to cover the following :
\begin{itemize}[leftmargin=*,noitemsep]
    \item Limitations - Discuss four key limitations with this work that we plan to address in our future studies.
    \item Dataset -  Discuss the crawled github data characteristics in detail
    \item Annotator Profile -  Discuss the demography and key details of annotators who helped prepare the study dataset
     \item Annotator Instruction -  Discuss in detail 
     the instructions and guidance provided to annotators
     \item Annotation Validation -  Discuss in detail the steps taken to review annotations
     \item Task Modelling using Baseline Models -  List all the prompts tried to get the most accurate functionalities
     \item Model Hyperparameters -  Key hyper-parameters used to reproduce results
     \item Quantitative Results -  Discuss results in detail for the different settings and thresholds 
\end{itemize}

\subsection{Limitations}
\label{limitations}
There are four major limitations in this work that could be addressed in future research. First, the study focused on $2101$ samples, there could be more unknown ways of describing functionalities that the current models may not be able to handle. This can be addressed by increasing the dataset size. Second, as shown in Figure \ref{fig:rating_matching}, we found human errors during the annotation process where, for a few samples, unwanted functionalities were added and some wanted functionalities were missed. But this can be handled by expanding the validation efforts to the rest of the samples. Third, handling very long {\readme} files is a challenge as we have a maximum of $2048$ token limit for models. There is promising research in this direction to support longer token limit. Fourth, defining the reference set of functionalities is sometimes an ill-posed problem because different humans may perceive the {\readme} differently and they may conceive the set of functionalities differently. But we hope to educate annotators by discussing more number of ground truth samples.

\subsection{Dataset}
\label{sec:Dataset}
\begin{figure}[!t]
  \centering
  \includegraphics[width=\linewidth]{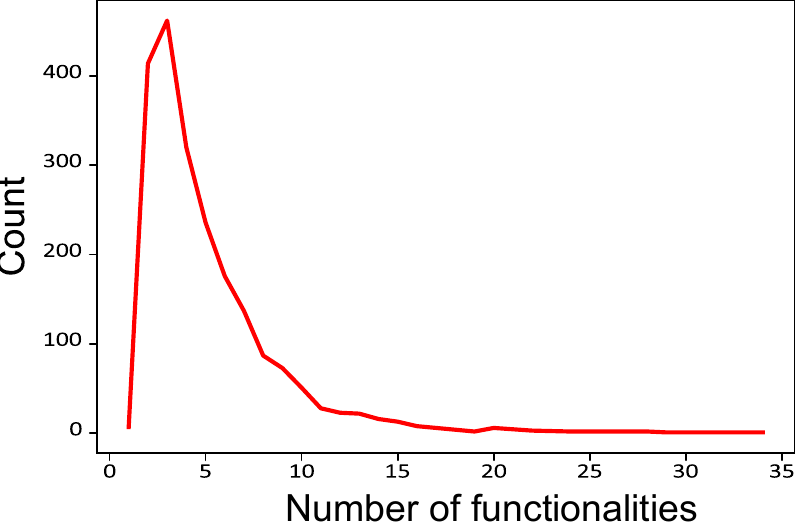}
 \caption{Functionalities count distribution of the {\FuncRead} dataset.}
  \label{fig:feature_stats}
\end{figure}

 Table \ref{table:license_stats} shows the license distribution for the $2101$.
 Figure \ref{fig:feature_stats} represents the functionalities count distribution for the repositories. {\readme} files. We plan to release this dataset post review period.
\begin{table}[t]

	\setlength\tabcolsep{12.5pt}
	\begin{tabularx}{\textwidth}{lrr}
 
		\cmidrule[\heavyrulewidth]{1-3}
  
		\thead{\bf License}  & {\centering \thead{\bf Count}} & {\centering \thead{\bf Count Percentage(\%)}} \\ %
		\cmidrule{1-3}
		
		MIT & $1436$  & $68.34$\\
      Apache & $334$ & $15.90$\\
      BSD & $325$ & $15.47$\\
      EPL & $6$ & $0.29$\\
		
		\cmidrule[\heavyrulewidth]{1-3}
	\end{tabularx}

\caption{License-wise split of {\FuncRead} dataset.}

\label{table:license_stats}

\end{table}

\subsection{Annotators Profile}
\label{sec:annotataor_profile}
To prepare the dataset, we requested participation from nine software engineers based out of Asia. The participants were identified based on their prior experience working on application modernization projects listed on their profile page. On an average, the participants had industrial experience of $13$ years in different software engineering roles. We requested seven participants to annotate the $2101$ different GitHub {\readme} files. Once extractive and abstractive functionalities were annotated, we employed 2 new participants to perform the verification step. We individually discussed the task details, expectations, the tentative average time that might be needed (~5 minutes per annotation), and the research goal and got their consensus before providing them with the annotation instruction.


\subsection{Annotation Instructions}
\label{sec:annotataor_instructions}
Following were the instructions given to the seven annotators : 
\textit{
\begin{itemize}
    \item We thank you for agreeing to annotate. An excel sheet will be given with the following information
        \begin{itemize}
            \item Repository id
            \item Readme URL
            \item Extractive functionalities
            \item Abstractive functionalities
        \end{itemize}
    \item First row will be filled for convenience.
    \item For each repository id two types of annotations are requested to be done
        \begin{itemize}
            \item Extractive: Copy and paste the functionalities as numbered lists.
            \item Abstractive: Write the functionality in your own words.
                \begin{itemize}
                    \item \textbf{NOTE:} Please do not copy-paste for this. Please try to be as descriptive as possible i.e., introduce new words to describe instead of reusing the same set of words.
                \end{itemize}
        \end{itemize}
    \item Please write/copy-paste each functionality in the new line as a numbered list.
    \item Please make sure that number of abstractive and extractive functionalities are the same.
    \item Few things to take care
        \begin{itemize}
            \item Do not include future/expected functionalities/roadmap/TODO/planned
            \item Please do not click on any link to find more functionalities. Whatever functionalities are present in the {\readme}, please include those only.
            \item Do not include application -- meaning what is possible with that functionality or repository.
            \item In Progress/partial functionalities can be included.
        \end{itemize}
\end{itemize}
}
All the annotators were given the same set of instructions so as to maintain consistency. Annotators' doubts were clarified on regular basis. The generated dataset was reviewed by the authors internal review board and was deemed suitable to be published for research.



\subsubsection{Annotator Validation Example}
Let us understand above ratings via an example.
For the {\readme} given in Figure \ref{fig:desc_func}, suppose following extractive functionalities were annotated by an annotator:
\begin{itemize} [leftmargin=*, noitemsep]
 \item  {\em allow users to login}
 \item  {\em lookup stock quotes}
 \item  {\em buy or sell stock shares}
 \item  {\em provides a real-world java EE workload}
\end{itemize}
It is now clear that the annotator in this specific case has missed one of the functionality, namely ``\textit{view their portfolio}'' and added an extra functionality namely ``{\em provides a real-world java EE workload}''. Therefore, a rating of $4$ would be assigned during the human validation step.

\subsection{Task Modelling using ChatGPT, Bard}
\label{sec:taskmodellinggpt}
To understand what prompts helps best to list the functionalities, we tried various prompt on ChatGPT and Bard baseline models. Some of them are as follows:
\begin{itemize}[leftmargin=*,noitemsep]
\item {\em List all the features for the above text.}
\item {\em List all the functionalities for the above text.}
\item {\em List all the features from above text. Each features should be in individual line without headings.}
\item {\em List all the features from above text. Each features should be in individual line without headings. Each features should be in individual line without headings.}
\item {\em List all the features from above text. Each features should be in individual line without headings. Each features should be in individual line without headings. Do not include features related to license}

\end{itemize}

\subsection{Evaluation Metrics}
To evaluate the quality of the generated functionalities, we align them to the gold annotated functionalities via bipartite matching. We perform three kinds of bipartite matching: i) one-to-one, ii) one-to-many, and iii) weighted one-to-many. 

In any of these bipartite graphs, we have model-generated functionalities as nodes on one side and gold (ground truth) functionalities as nodes on the other side. The presence or absence of an edge in this bipartite graph is decided by the similarity scores between the corresponding sentences. Figure \ref{fig:matching} captures an illustration. For computing the similarity score, we used SentenceTransformer\footnote{\url{https://www.sbert.net/}} and generated the sentence embeddings for both model-generated and gold functionalities sentences. Next, we computed a cosine similarity between these two vectors, and experimented with multiple thresholds to decide whether the edge should be present in the bipartite graph. In our experiments we found threshold $0.3$ matches the most with the human judgment. 
A lower threshold was giving poor-quality mapping with excessively matched pairs. A higher value was giving high-quality mapping but the number of matched pairs was very less. We used the maximum\_bipartite\_matching\footnote{\url{https://docs.scipy.org/doc/scipy/reference/generated/scipy.sparse.csgraph.maximum_bipartite_matching.html}} function from SciPy library to perform the maximum (weighted or unweighted) bipartite matching. Based on the matched pairs, we compute Precision ($P$), Recall ($R$), and $F_1$ scores to measure the generation capability. 

For fine-tuning the models, we used extractive functionalities as gold, and because of it, we employed ROUGE-$1$, ROUGE-$2$, ROUGE-L scores to check the lexical matching quality of generated functionalities at an individual level. Since all the considered models are generative models, there is a high chance that it would introduce new tokens while generating functionalities. Hence, we also used BERTScore \cite{zhang2019bertscore} to capture the semantic similarity between the matched pairs.

After analyzing the generated functionalities, we realized that the model sometimes combines multiple functionalities into a single generated sentence (see Figure \ref{fig:matching}). Therefore, there is a need for many-to-one bipartite matching where multiple gold functionalities are allowed to map into a single generated functionality. There are two kinds of results we show in many-to-one bipartite matching. The first one is {\em many-to-one} $P$, $R$, and $F_1$ scores, where all the edges in the bipartite matching are given a score of $1$. The second is {\em weighted many-to-one} $P$, $R$, and $F_1$ scores, where for each of the model-generated functionality that is matched with multiple gold functionalities, each matched edge is assigned a weight that is inversely proportional to the number of functionalities matched. We take the reciprocal of the number of matched edges and assign that as a weight to all the incoming edges for that particular model-generated functionality. For example, consider the third functionality sentence generated by the model in Figure \ref{fig:matching}, which reads \textit{``It also has ranking functions based on peewee, utility function to add FTS5 auxiliary functions and an FTS5 aux function implementation.''} Now, each matched edge incident on this node gets a weight of $1/3$ for weighted many-to-one bipartite matching.

\begin{table*}[!ht]
	\setlength\tabcolsep{8.5pt}
	\begin{tabularx}{\textwidth}{crrrrrrrrr}
 
		\toprule
        
        \multirow{2}{*}{\thead{\bf Model}}  & \multicolumn{3}{c}{\thead{\bf ROUGE-1}} & \multicolumn{3}{c}{\thead{\bf ROUGE-2}} & \multicolumn{3}{c}{\thead{\bf ROUGE-L}} \\
        \cmidrule(l{5mm}r{5mm}){2-4}  \cmidrule(l{5mm}r{5mm}){5-7} \cmidrule(l{5mm}r{5mm}){8-10}
		  & \thead{\bm {$F_1$}} & \thead{\bf P} & \thead{\bf R} &\thead{\bm {$F_1$}}  & \thead{\bf P} & \thead{\bf R} &\thead{\bm {$F_1$}}  & \thead{\bf P} & \thead{\bf R} \\ %
		\midrule
		
        ChatGPT & 0.607 &  0.576 & 0.792 & 0.467 & 0.448 & 0.604 & 0.589 & 0.558 & 0.772		\\ 
        Bard & 0.687 & 0.719 & 0.764 & 0.583 & 0.617 & 0.636 & 0.681 & 0.711 & 0.758 \\ 
        StarCoderbase-1b & 0.765 & 0.752 & 0.868 & 0.677 & 0.664 & 0.772 & 0.763 & 0.750 & 0.864	\\
        StarCoderbase-7b & 0.742 & 0.766 & 0.813 & 0.626 & 0.639 & 0.688 & 0.739 & 0.762 & 0.809		\\
        Phi-2 & 0.664 & 0.667 & 0.775 & 0.567 & 0.567 & 0.662 & 0.661 & 0.663 & 0.769 \\
        Llama2-7b & 0.734 & 0.762 & 0.806 & 0.637 & 0.655 & 0.699 & 0.732 & 0.758 & 0.802 \\
        CodeLlama-7b & 0.772 & 0.797 & 0.833 & 0.681 & 0.699 & 0.735 & 0.770 & 0.795 & 0.830	\\
        \bottomrule
	\end{tabularx}

\caption{Results for many-to-one matched pairs with threshold = $0.3$.
}

\label{table:results_many2one_lexical_0.3}

\end{table*}

\begin{table}[!ht]
	\setlength\tabcolsep{9pt}
	\begin{tabularx}{\textwidth}{rrrr}
 
		\cmidrule[\heavyrulewidth]{1-4}

        \multirow{2}{*}{\thead{\bf Model}}  & \multicolumn{3}{c}{\thead{\bf BERTScore}}\\
        \cmidrule(l{5mm}r{5mm}){2-4}
		   & \thead{ $\bm F_1$ } & \thead{\bf P} & \thead{\bf R} \\ %
		\cmidrule{1-4}
		
         ChatGPT & 0.918 & 0.909 & 0.929	\\ 
        Bard & 0.920 & 0.917 & 0.924 \\ 
        StarCoderbase-1b & 0.950 & 0.944 & \textbf{0.958}	\\
        StarCoderbase-7b & 0.941 & 0.940 & 0.944	\\
        Phi-2 & 0.935 & 0.931 & 0.941	\\
        Llama2-7b & 0.941 & 0.938 & 0.945 \\
		CodeLlama-7b & \textbf{0.951} & \textbf{0.950} & 0.953 \\
		\cmidrule[\heavyrulewidth]{1-4}
	\end{tabularx}

\caption{Results for many-to-one matched pairs with threshold = $0.3$.
}

\label{table:results_many2one_semantic_0.3}

\end{table}


\begin{table*}[!ht]
\setlength\tabcolsep{8.5pt}
	\setlength\tabcolsep{10.5pt}
	\begin{tabularx}{\textwidth}{crrrrrrrr}
 
		\toprule
  
		\thead{\bf Model} & \thead{\bm {$F_1^\#$}} &\thead{\bf $P^\#$}  & \thead{\bf $R^\#$} &\thead{\bm {$F_1^*$}}  & \thead{\bf $P^*$} & \thead{\bf $R^*$} &\thead{\bm {$F_1^+$}}  & \thead{\bf $P^+$} \\ %
		\midrule
		
        ChatGPT & 0.431 & 0.314 & 0.849 & 0.415 & 0.293 & 0.878 & 0.395 & 0.276  \\ 
        Bard & 0.614 & 0.575 & 0.753 & 0.619 & 0.556 & 0.795 & 0.594 & 0.522	 \\ 
        StarCoderbase-1b & 0.738 & 0.778 & 0.752 & 0.771 & 0.767 & 0.819 & 0.735 & 0.712	\\
        StarCoderbase-7b & 0.713 & 0.764 & 0.723 & 0.745 & 0.754 & 0.783 & 0.713 & 0.701	\\
        Phi- 2 & 0.213 & 0.158 & 0.604 & 0.211 & 0.152 & 0.661 & 0.200 & 0.143\\
        Llama2-7b & 0.653 & 0.697 & 0.669 & 0.669 & 0.671 & 0.726 & 0.633 & 0.623	 \\
        CodeLlama-7b & 0.752 & 0.792 & 0.761 & 0.777 & 0.780 & 0.816 & 0.750 & 0.737 \\
        \bottomrule
	\end{tabularx}

\caption{Result comparison for various fine-tuned models against out-of-the box large models for threshold = $0.4$. \# represents one-to-one bipartite matching, * represents many-to-one bipartite matching, + represents weighted many-to-one  bipartite matching. 
}

\label{table:results_0.4}

\end{table*}

\begin{table*}[!ht]
	\setlength\tabcolsep{8.5pt}
	\begin{tabularx}{\textwidth}{crrrrrrrrr}
 
		\toprule
        
        \multirow{2}{*}{\thead{\bf Model}}  & \multicolumn{3}{c}{\thead{\bf ROUGE-1}} & \multicolumn{3}{c}{\thead{\bf ROUGE-2}} & \multicolumn{3}{c}{\thead{\bf ROUGE-L}} \\
        \cmidrule(l{5mm}r{5mm}){2-4}  \cmidrule(l{5mm}r{5mm}){5-7} \cmidrule(l{5mm}r{5mm}){8-10}
		  & \thead{\bm {$F_1$}} & \thead{\bf P} & \thead{\bf R} &\thead{\bm {$F_1$}}  & \thead{\bf P} & \thead{\bf R} &\thead{\bm {$F_1$}}  & \thead{\bf P} & \thead{\bf R} \\ %
		\midrule
		
        ChatGPT & 0.527 & 0.509 & 0.670 & 0.391 & 0.381 & 0.489 & 0.512 & 0.493 & 0.652		\\ 
        Bard & 0.701 & 0.734 & 0.764 & 0.590 & 0.621 & 0.628 & 0.694 & 0.725 & 0.756 \\ 
        StarCoderbase-1b & 0.813 & 0.804 & 0.903 & 0.721 & 0.713 & 0.805 & 0.811 & 0.801 & 0.899 \\
        StarCoderbase-7b & 0.820 & 0.848 & 0.869 & 0.696 & 0.715 & 0.744 & 0.818 & 0.845 & 0.867 \\
        Phi-2 & 0.733 & 0.741 & 0.831 & 0.631 & 0.635 & 0.720 & 0.730 & 0.736 & 0.826 \\
        Llama2-7b & 0.812 & 0.842 & 0.863 & 0.714 & 0.739 & 0.757 & 0.809 & 0.838 & 0.858 \\
        CodeLlama-7b & 0.834 & 0.858 & 0.880 & 0.737 & 0.758 & 0.778 & 0.832 & 0.855 & 0.878	\\
        \bottomrule
	\end{tabularx}

\caption{Results for one-to-one matched pairs with threshold = $0.4$.
}

\label{table:results_one2one_lexical_0.4}

\end{table*}

\begin{table}[!ht]
	\setlength\tabcolsep{9pt}
	\begin{tabularx}{\textwidth}{rrrr}
 
		\cmidrule[\heavyrulewidth]{1-4}

        \multirow{2}{*}{\thead{\bf Model}}  & \multicolumn{3}{c}{\thead{\bf BERTScore}}\\
        \cmidrule(l{5mm}r{5mm}){2-4}
		   & \thead{ $\bm F_1$ } & \thead{\bf P} & \thead{\bf R} \\ %
		\cmidrule{1-4}
		
         ChatGPT & 0.906 & 0.901 & 0.913	\\ 
        Bard & 0.923 & 0.919 & 0.927 \\ 
        StarCoderbase-1b & 0.951 & 0.945 & 0.959	\\
        StarCoderbase-7b & 0.946 & 0.944 & 0.949 \\
        Phi-2 & 0.940 & 0.937 & 0.944	\\
        Llama2-7b & 0.946 & 0.943 & 0.950 \\
		CodeLlama-7b & 0.948 & 0.947 & 0.950 \\
		\cmidrule[\heavyrulewidth]{1-4}
	\end{tabularx}

\caption{Results for one-to-one matched pairs with threshold = $0.4$.
}

\label{table:results_one2one_semantic_0.4}

\end{table}


\begin{table*}[!ht]
	\setlength\tabcolsep{8.5pt}
	\begin{tabularx}{\textwidth}{crrrrrrrrr}
 
		\toprule
        
        \multirow{2}{*}{\thead{\bf Model}}  & \multicolumn{3}{c}{\thead{\bf ROUGE-1}} & \multicolumn{3}{c}{\thead{\bf ROUGE-2}} & \multicolumn{3}{c}{\thead{\bf ROUGE-L}} \\
        \cmidrule(l{5mm}r{5mm}){2-4}  \cmidrule(l{5mm}r{5mm}){5-7} \cmidrule(l{5mm}r{5mm}){8-10}
		  & \thead{\bm {$F_1$}} & \thead{\bf P} & \thead{\bf R} &\thead{\bm {$F_1$}}  & \thead{\bf P} & \thead{\bf R} &\thead{\bm {$F_1$}}  & \thead{\bf P} & \thead{\bf R} \\ %
		\midrule
		
        ChatGPT & 0.632 & 0.605 & 0.799 & 0.493 & 0.476 & 0.625 & 0.616 & 0.588 & 0.781	\\ 
        Bard & 0.740 & 0.768 & 0.813 & 0.639 & 0.667 & 0.692 & 0.735 & 0.760 & 0.807 \\ 
        StarCoderbase-1b & 0.810 & 0.796 & 0.909 & 0.724 & 0.710 & 0.823 & 0.808 & 0.794 & 0.906	\\
        StarCoderbase-7b & 0.805 & 0.824 & 0.868 & 0.688 & 0.699 & 0.750 & 0.802 & 0.820 & 0.865		\\
        Phi-2 & 0.739 & 0.738 & 0.845 & 0.644 & 0.642 & 0.744 & 0.735 & 0.734 & 0.839		\\
        Llama2-7b & 0.793 & 0.818 & 0.855 & 0.697 & 0.716 & 0.755 & 0.791 & 0.815 & 0.851 \\
        CodeLlama-7b & 0.828 & 0.847 & 0.883 & 0.738 & 0.754 & 0.790 & 0.826 & 0.845 & 0.881	\\
        \bottomrule
	\end{tabularx}

\caption{Results for many-to-one matched pairs with threshold = $0.4$.
}

\label{table:results_many2one_lexical_0.4}

\end{table*}

\begin{table}[!ht]
	\setlength\tabcolsep{9pt}
	\begin{tabularx}{\textwidth}{rrrr}
 
		\cmidrule[\heavyrulewidth]{1-4}

        \multirow{2}{*}{\thead{\bf Model}}  & \multicolumn{3}{c}{\thead{\bf BERTScore}}\\
        \cmidrule(l{5mm}r{5mm}){2-4}
		   & \thead{ $\bm F_1$ } & \thead{\bf P} & \thead{\bf R} \\ %
		\cmidrule{1-4}
		
         ChatGPT & 0.921 & 0.912 & 	0.930	\\ 
        Bard & 0.925 & 0.922 & 0.930 \\ 
        StarCoderbase-1b & 0.955 & 0.948 & 0.963	\\
        StarCoderbase-7b & 0.947 & 0.946 & 0.950 \\
        Phi-2 & 0.947 & 0.943 & 0.952 \\
        Llama2-7b & 0.946 & 0.943 & 0.951 \\
		CodeLlama-7b & 0.953 & 0.952 & 0.955 \\
		\cmidrule[\heavyrulewidth]{1-4}
	\end{tabularx}

\caption{Results for many-to-one matched pairs with threshold = $0.4$.
}

\label{table:results_many2one_semantic_0.4}

\end{table}

\begin{table*}[!ht]
\setlength\tabcolsep{8.5pt}
	\setlength\tabcolsep{10.5pt}
	\begin{tabularx}{\textwidth}{crrrrrrrr}
 
		\toprule
  
		\thead{\bf Model} & \thead{\bm {$F_1^\#$}} &\thead{\bf $P^\#$}  & \thead{\bf $R^\#$} &\thead{\bm {$F_1^*$}}  & \thead{\bf $P^*$} & \thead{\bf $R^*$} &\thead{\bm {$F_1^+$}}  & \thead{\bf $P^+$} \\ %
		\midrule
		
        ChatGPT & 0.398 & 0.290 & 0.783 & 0.392 & 0.280 & 0.806 & 0.380 & 0.269  \\ 
        Bard & 0.553 & 0.520 & 0.672 & 0.562 & 0.514 & 0.702 & 0.547 & 0.492 \\ 
        StarCoderbase-1b & 0.710 & 0.747 & 0.724 & 0.730 & 0.743 & 0.763 & 0.711 & 0.712	\\
        StarCoderbase-7b & 0.682 & 0.731 & 0.689 & 0.702 & 0.726 & 0.724 & 0.685 & 0.697		\\
        Phi- 2 & 0.198 & 0.148 & 0.558 & 0.199 & 0.145 & 0.593 & 0.192 & 0.139 \\
        Llama2-7b & 0.611 & 0.647 & 0.624 & 0.621 & 0.634 & 0.656 & 0.602 & 0.608		 \\
        CodeLlama-7b & 0.726 & 0.756 & 0.735 & 0.742 & 0.7506 & 0.769 & 0.726 & 0.723 \\
        \bottomrule
	\end{tabularx}

\caption{Result comparison for various fine-tuned models against out-of-the box large models for threshold = $0.5$. \# represents one-to-one bipartite matching, * represents many-to-one bipartite matching, + represents weighted many-to-one  bipartite matching. 
}

\label{table:results_0.5}

\end{table*}
\begin{table*}[!ht]
	\setlength\tabcolsep{8.5pt}
	\begin{tabularx}{\textwidth}{crrrrrrrrr}
 
		\toprule
        
        \multirow{2}{*}{\thead{\bf Model}}  & \multicolumn{3}{c}{\thead{\bf ROUGE-1}} & \multicolumn{3}{c}{\thead{\bf ROUGE-2}} & \multicolumn{3}{c}{\thead{\bf ROUGE-L}} \\
        \cmidrule(l{5mm}r{5mm}){2-4}  \cmidrule(l{5mm}r{5mm}){5-7} \cmidrule(l{5mm}r{5mm}){8-10}
		  & \thead{\bm {$F_1$}} & \thead{\bf P} & \thead{\bf R} &\thead{\bm {$F_1$}}  & \thead{\bf P} & \thead{\bf R} &\thead{\bm {$F_1$}}  & \thead{\bf P} & \thead{\bf R} \\ %
		\midrule
		
        ChatGPT & 0.632 & 0.617 & 0.752 & 0.499 & 0.488 & 0.611 & 0.617 & 0.602 & 0.736		\\ 
        Bard & 0.796 & 0.822 & 0.843 & 	0.696 & 0.721 & 0.739 & 0.788 & 0.812 & 0.835	 \\ 
        StarCoderbase-1b & 0.866 & 0.858 & 0.943 & 0.796 & 0.790 & 0.876	& 0.864 & 0.855 & 0.941	\\
        StarCoderbase-7b & 0.850 & 0.875 & 0.896 & 0.743 & 0.759 & 0.795 & 0.849 & 0.872 & 0.895		\\
        Phi-2 & 0.800 & 0.806 & 0.882 & 0.718 & 0.725 & 0.797 & 0.799 & 0.805 & 0.878	\\
        Llama2-7b & 0.858 & 0.889 & 0.905 & 0.784 & 0.813 & 0.834 & 0.855 & 0.886 & 0.902 \\
        CodeLlama-7b & 0.881 & 0.901 & 0.920 & 0.791 & 0.813 & 0.834 & 0.880 & 0.899 & 0.919 \\
        \bottomrule
	\end{tabularx}

\caption{Results for one-to-one matched pairs with threshold = $0.5$.
}

\label{table:results_one2one_lexical_0.5}

\end{table*}

\begin{table}[!ht]
	\setlength\tabcolsep{9pt}
	\begin{tabularx}{\textwidth}{rrrr}
 
		\cmidrule[\heavyrulewidth]{1-4}

        \multirow{2}{*}{\thead{\bf Model}}  & \multicolumn{3}{c}{\thead{\bf BERTScore}}\\
        \cmidrule(l{5mm}r{5mm}){2-4}
		   & \thead{ $\bm F_1$ } & \thead{\bf P} & \thead{\bf R} \\ %
		\cmidrule{1-4}
		
         ChatGPT & 0.920 & 0.914 & 0.928\\ 
        Bard & 0.937 & 0.934 & 0.941 \\ 
        StarCoderbase-1b & 0.962 & 0.956 & 0.969	\\
        StarCoderbase-7b & 0.954 & 0.953 & 0.956 \\
        Phi-2 & 0.956 & 0.953 & 0.959 \\
        Llama2-7b & 0.954 & 0.953 & 0.956 \\
		CodeLlama-7b & 0.959 & 0.959 & 0.961 \\
		\cmidrule[\heavyrulewidth]{1-4}
	\end{tabularx}

\caption{Results for one-to-one matched pairs with threshold = $0.5$.
}

\label{table:results_one2one_semantic_0.5}

\end{table}

\begin{table*}[!ht]
	\setlength\tabcolsep{8.5pt}
	\begin{tabularx}{\textwidth}{crrrrrrrrr}
 
		\toprule
        
        \multirow{2}{*}{\thead{\bf Model}}  & \multicolumn{3}{c}{\thead{\bf ROUGE-1}} & \multicolumn{3}{c}{\thead{\bf ROUGE-2}} & \multicolumn{3}{c}{\thead{\bf ROUGE-L}} \\
        \cmidrule(l{5mm}r{5mm}){2-4}  \cmidrule(l{5mm}r{5mm}){5-7} \cmidrule(l{5mm}r{5mm}){8-10}
		  & \thead{\bm {$F_1$}} & \thead{\bf P} & \thead{\bf R} &\thead{\bm {$F_1$}}  & \thead{\bf P} & \thead{\bf R} &\thead{\bm {$F_1$}}  & \thead{\bf P} & \thead{\bf R} \\ %
		\midrule
		
        ChatGPT & 0.676 & 0.653 & 0.811 & 0.545 & 0.527 & 0.671 & 0.662 & 0.638 & 0.794	\\ 
        Bard & 0.809 & 0.827 & 0.869 & 0.718 & 0.736 & 0.777 & 0.804 & 0.820 & 0.863 \\ 
        StarCoderbase-1b & 0.841 & 0.829 & 0.929 & 0.770 & 0.758 & 0.859 & 0.840 & 0.826 & 0.925 \\
        StarCoderbase-7b & 0.837 & 0.855 & 0.895 & 0.731 & 0.742 & 0.793 & 0.835 & 0.852 & 0.892			\\
        Phi-2 & 0.791 & 0.792 & 0.882 & 0.709 & 0.710 & 0.801 & 0.787 & 0.788 & 0.877 \\
        Llama2-7b & 0.831 & 0.857 & 0.887 & 0.754 & 0.778 & 0.811 & 0.828 & 0.854 & 0.883	 \\
        CodeLlama-7b & 0.870 & 0.886 & 0.917 & 0.781 & 0.800 & 	0.833 & 0.868 & 0.885 & 0.915\\
        \bottomrule
	\end{tabularx}

\caption{Results for many-to-one matched pairs with threshold = $0.5$.
}

\label{table:results_many2one_lexical_0.5}

\end{table*}

\begin{table}[!ht]
	\setlength\tabcolsep{9pt}
	\begin{tabularx}{\textwidth}{rrrr}
 
		\cmidrule[\heavyrulewidth]{1-4}

        \multirow{2}{*}{\thead{\bf Model}}  & \multicolumn{3}{c}{\thead{\bf BERTScore}}\\
        \cmidrule(l{5mm}r{5mm}){2-4}
		   & \thead{ $\bm F_1$ } & \thead{\bf P} & \thead{\bf R} \\ %
		\cmidrule{1-4}
		
         ChatGPT & 0.928 & 0.919 & 0.937	\\ 
        Bard & 0.938 & 0.936 & 0.942 \\ 
        StarCoderbase-1b & 0.962 & 0.956 & 0.969\\
        StarCoderbase-7b & 0.954 & 0.953 & 0.955	\\
        Phi-2 & 0.958 & 0.956 & 0.962	\\
        Llama2-7b & 0.953 & 0.951 & 0.956 \\
		CodeLlama-7b & 0.953 & 0.951 & 0.956 \\
		\cmidrule[\heavyrulewidth]{1-4}
	\end{tabularx}

\caption{Results for many-to-one matched pairs with threshold = $0.5$.
}

\label{table:results_many2one_semantic_0.5}

\end{table}




\subsection{Model Hyperparameters}
Table \ref{table:hyperparams} shows the important hyperparamters that can be used to reproduce results. Rest of the hyperparamters are the default ones present in Huggingface Trainer API.

\begin{table*}[t]

	\setlength\tabcolsep{12.5pt}
	\begin{tabularx}{\textwidth}{rrrrrr}
 
		\cmidrule[\heavyrulewidth]{1-6}
  
		\thead{\bf Model}  & {\centering \thead{\bf Learning Rate}}& {\centering \thead{\bf Learning Rate Scheduler}} & {\centering \thead{\bf Batch Size}}  & {\centering \thead{\bf Step Size}}& {\centering \thead{\bf Epochs}} \\ %
		\cmidrule{1-6}
		
        StarCoderbase-1b & 5e-7 & cosine & 2 & 100 & 10	\\
        StarCoderbase-7b & 5e-6 & cosine & 1 & 100 & 5	\\
        Phi-2 & 5e-7 & cosine & 1 & 100 & 10	\\
        Llama2-7b & 5e-6 & cosine & 1 & 100 & 5	\\
		CodeLlama-7b & 5e-5 & cosine & 1 & 100 & 5	 \\
		\cmidrule[\heavyrulewidth]{1-6}
	\end{tabularx}

\caption{Hyperparamaters for the different fine-tuned models}

\label{table:hyperparams}

\end{table*}

\subsection{Quantitative Results}
All experiments were performed on an $A100$ $80GB$ GPU machine. 

We report results on the discussed metrics for all the fine-tuned models and compare them against the ChatGPT and Bard. 
Table \ref{table:results_1_0.3} shows the $P$, $R$, and $F_1$ scores for the three bipartite matching schemes. We do not report $R$ for weighted many-to-one bipartite matching as it is the same as $R$ for many-to-one bipartite matching. Results in tables \ref{table:results_1_0.3}, \ref{table:results_one2one_lexical_0.3}, and \ref{table:results_one2one_semantic_0.3}, are restricted over that subset of test samples for which each of these models outputs a nonempty string and also yields at least one matched pair during the bipartite matching procedure. The total comparable test samples thus came down to $69$.

From table \ref{table:results_1_0.3}, we can observe that all the fine-tuned models significantly outperform ChatGPT and Bard across $P$, $R$, and $F_1$ measures. We can see that the $F_1$ score of one-to-one bipartite matching for ChatGPT is $0.459$ and for Bard is $0.653$ which are much smaller as compared to code models. Table \ref{table:results_one2one_lexical_0.3} further shows the ROUGE scores for one-to-one matched pairs. Again we see that the functionalities generated by the fine-tuned models have a relatively higher lexical similarity. Table \ref{table:results_one2one_semantic_0.3} shows BERTScore which is again higher than ChatGPT and Bard.
Tables \ref{table:results_many2one_lexical_0.3} and \ref{table:results_many2one_semantic_0.3} shows many-to-one results for threshold = 0.3.
The rest of the tables show results for other threshold values $0.4$ and $0.5$ and matching schemes. Count of common test samples across various models which have non-empty generations and have at least one matched pair are 85 and 98 for threshold values $0.4$ and $0.5$ respectively. An increase in ROUGE and BERTScore gives the illusion that a higher threshold value should be preferred but as mentioned earlier the number of functionalities generated/classified decreases too which is not much helpful as we lose out on many functionalities.
We recorded the responses from ChatGPT and Bard on November 25, 2023 for our experiments.


For the different task types and for threshold 0.4, please refer tables \ref{table:results_0.4}-\ref{table:results_many2one_semantic_0.4}.
For threshold 0.5, please refer tables \ref{table:results_0.5}-\ref{table:results_many2one_semantic_0.5}.





\end{document}